# ALBERT with Knowledge Graph Encoder Utilizing Semantic Similarity for Commonsense Question Answering

Byeongmin Choi[1], YongHyun Lee[1], Yeunwoong Kyung[2] and Eunchan Kim[3,*]

[1]Department of Computer Science and Engineering, Seoul National University, Seoul, 08826, Korea
[2]School of Computer Engineering, Hanshin University, Osan, 18101, Korea
[3]Department of Intelligence and Information, Seoul National University, Seoul, 08826, Korea
*Corresponding Author: Eunchan Kim. Email: eunchan@snu.ac.kr

**Abstract:** Recently, pre-trained language representation models such as bidirectional encoder representations from transformers (BERT) have been performing well in commonsense question answering (CSQA). However, there is a problem that the models do not directly use explicit information of knowledge sources existing outside. To augment this, additional methods such as knowledge-aware graph network (KagNet) and multi-hop graph relation network (MHGRN) have been proposed. In this study, we propose to use the latest pre-trained language model a lite bidirectional encoder representations from transformers (ALBERT) with knowledge graph information extraction technique. We also propose to applying the novel method, schema graph expansion to recent language models. Then, we analyze the effect of applying knowledge graph-based knowledge extraction techniques to recent pre-trained language models and confirm that schema graph expansion is effective in some extent. Furthermore, we show that our proposed model can achieve better performance than existing KagNet and MHGRN models in CommonsenseQA dataset.

## 1 Introduction

The recent deep learning revolution has triggered the generalization of artificial intelligence (AI) technology and is being used in various places in everyday life. This is because, compared to the past, deep learning technology has led to the development of artificial intelligence in many computer-related fields, such as computer vision, voice recognition, autonomous driving, natural language processing (NLP) and image recognition. One of the areas with the most notable recent advances is NLP. For example, bidirectional encoder representations from transformers (BERT), a representative model of recent NLP, outperforms among existing models used in various NLP tasks.

Many AI systems perform beyond humans for certain problems or similar problems that they have learned, but they rarely respond to completely new problems. Typical AI systems only use training data to learn models by reducing training loss but they do not have knowledge other than learning data. On the other hand, when human encounters a certain problem in their daily lives, they can solve the problem by using previous experience or knowledge acquired from the outside. Commonsense reasoning is the field of studying artificial intelligence systems that solve problems using accumulated external knowledge just like humans do.

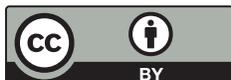

The commonsense question answering (CSQA) is a type of commonsense reasoning and a task that requires knowledge outside the problem to answer for a given question [1]. In the case of machine reading comprehension, which is a widely known question answering (QA) task, all information necessary for reasoning exists within the text of the problem [2]. However, CSQA requires information other than the question text, so it can be regarded as a more difficult problem. Consequently, for the CSQA problem, we need to build a model that can utilize external knowledge. For example, when we have a question 'what tools do the police use to arrest criminals?', then we are given 4 examples of handcuffs, carrots, cars, and rice. We do not know the answer from the question but we are able to know that handcuffs are the answer using our knowledge.

There are two main ways how a neural network model utilizes knowledge. First, the model stores knowledge inside the neural network during the learning. Pre-trained language models such as BERT [3] belong to this category. The second way is to infer using external knowledge in an explicit way. For example, you can refer to relevant sentences from large-scale textual information such as Wikipedia, and then use this as external knowledge. In addition, information can be directly acquired using a knowledge graph such as ConceptNet [4] and then we can do inference.

In case of pre-trained language model, it implicitly stores knowledge contained in large corpus inside a neural network. According to recent studies, these pre-trained language models alone can achieve sufficient performance for CSQA problems [5,6]. However, if we only use the pre-trained language models, there is a problem that we cannot use the knowledge that exists outside the models. Humankind has recorded knowledge and solved problems with this accumulated knowledge. If a similar method is applied to AIs, we can obtain better results. From this point of view, artificial neural network-based methods such as knowledge-aware graph network (KagNet) [5] and multi-hop graph relation network (MHGRN) [6] that utilize external knowledge graphs in QA tasks have been proposed.

In this study, in order to solve the CSQA problem, we apply the existing methods which use an external knowledge graph to the latest pre-trained language models. The contribution of this paper is as follows. (1) In the previous study, BERT or robustly optimized pretraining approach (RoBERTa) was used as a pre-trained language model but we propose to use the latest ALBERT as a pre-trained language model. (2) We also present a new methodology called schema graph expansion (SGE), which utilizes knowledge graphs more efficiently than previous studies. An extracted graph from external knowledge graphs to get relevant knowledge is called schema graph, which is a concept used in KagNet [5]. Existing knowledge graph-based techniques such as KagNet [5] and MHGRN [6] have limitation such that extracting and using only the concepts that appear in the problem. In this study, we increase flexibility by utilizing SGE which uses other concepts similar to those that appeared in the problem in the knowledge graph. (3) We conduct various experiments and analyze the results. Also, we show that our method proposed in this study can achieve better performance than conventional methods.

## 2 Background

### 2.1 Knowledge Graph

If we want an AI system to perform interactions with humans in general situations, rather than only in a limited range of special tasks, the system must have a level of commonsense comparable to that of humans. Knowledge representation is one of the ways in which the system stores and utilizes real-world knowledge in an appropriate form. A knowledge graph is a knowledge base that expresses real-world knowledge in a graph form. Knowledge graphs express knowledge using entities (concepts) and relationships. In a knowledge graph, we use 'triple' to refer to the smallest knowledge unit and it consists of two entities and one relationship which connects these two entities. Multiple triples are gathered to represent a knowledge graph. Fig. 1 shows an example of a knowledge graph. The words in the square are the entities and the line connecting the two entities is the relationship between them. For example, "tree" and "bird" are entities and they are connected by a relationship called "LocatedAt" which means that birds are located in trees.

**Figure 1:** Example of knowledge graphs

Representative knowledge graphs are ConceptNet, Freebase, and DBpedia and the model proposed in this study uses ConceptNet among them. The number of entities in ConceptNet is approximately 8 million, and the number of edges connecting entities is about 21 million. Since all the questions in the CommonsenseQA dataset are all written in English, only English vocabulary from ConceptNet was extracted and used in this study. There are about 1.5 million English vocabulary words in ConceptNet. ConceptNet provides an interface that allows you to search directly through website as well as through application programming interface (API). For example, Fig. 2 is the result of a query which is related to the vocabulary "computer".

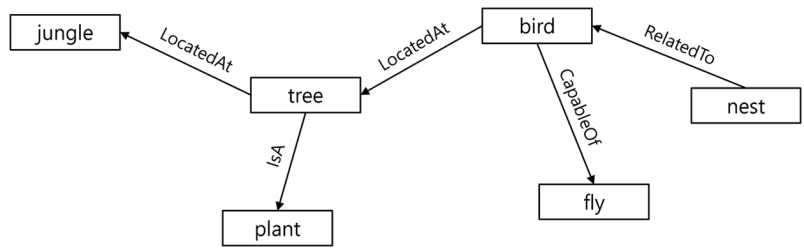

**Figure 2:** Result of searching "computer" from ConceptNet website

## 2.2 Natural Language Processing (NLP)

NLP is a field of study about how a computer analyzes and processes human's natural language. This means that the computer understands and processes the meaning of natural language sentences used by humans. However, since most of AI systems perform all operations through vectors, all natural language words such as nouns and verbs must be converted into vectors of fixed length. This process is called embedding and it is the most basic step in a NLP method that utilize artificial neural networks.

Before embedding a sentence, we need to convert a token, such as a word, into a vector representation, a smaller unit that makes up the sentence. Various word embedding methods such as word2vec [7], global vector (GloVe) [8], and fastText [9] have been proposed and are widely used in various fields of NLP. These methods can solve the problem that natural language word representation has no symbolic similarity despite its semantic similarity, which can occur in one-hot-encoding. For example, the words "king" and "queen" have very similar meanings in terms of representing the head of a country. However, we cannot find any similarity in the pattern of letters that make up the word. Word embedding methods allow you to preserve semantic similarities between words to be represented in numerical ways by learning so that similar or highly relevant words become similar vectors.

The simplest way to obtain sentence embeddings from word embeddings is to average the embedding vectors of the words that make up the sentence and use them as embeddings in the sentence. However, this method has a critical problem that two sentences get the same sentence embedding as long as the two sentences are composed of the same words. In other words, there is a problem that it does not reflect the order of arrangement of words. A better way to obtain a sentence representation is to use recurrent neural network (RNN), a kind of deep learning model [10]. A RNN is a neural network that uses not only the input of the current stage but also the hidden state of the stage. Thus, it can reflect the order of words that make up a sentence. The structure of the RNN model is shown in Fig. 3. The internal state of the RNN, changed by the word entered in the very previous step (t−1), is used again as the input with the word (t) in the current step.

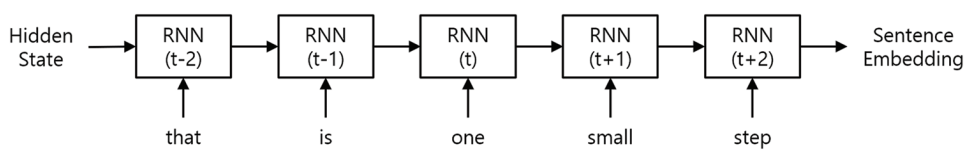

**Figure 3:** Example of recurrent neural network(RNN) structure

As such, embeddings play a very fundamental and essential role in NLP. Various studies have been conducted to obtain good embeddings. Recently, a pre-trained language representation model represented by BERT has emerged. A pre-trained language model is an application of transfer learning techniques [11] to the field of NLP. This is a technique to pre-train language models for large corpus. It has been proved experimentally that word embeddings generated by these pre-trained language models perform well in most NLP tasks [3]. Also, there are studies using lexicalized dependency paths for relation extraction in the NLP field [12,13].

The most famous pre-trained language model is BERT [3]. It was built using the encoder structure of Transformer [14] which is the latest machine translation model. At the time of publication, BERT proved its usefulness by achieving the highest performance in all tasks of the general language understanding evaluation (GLUE) benchmark [15]. GLUE benchmark is a benchmark to test the general performance of NLP models.

Although pre-trained language models show quite good performance in CSQA, various methods were proposed that utilize additional external knowledge such as ConceptNet to enhance models' interpretability.

KagNet is a method to extract a knowledge graph related to a CSQA problem from the external knowledge graph. KagNet consists of three neural network modules: the first module calculates node embedding of a graph through a graph convolution neural network. The second module calculates embedding for paths in the graph through long short-term memory (LSTM). The final module calculates the embedding for the entire graph with the attention mechanism to the path embeddings [5].

MHGRN was proposed to solve the problem that the processing time of embedding the graph based on the set of paths increases as the size of the knowledge graph increases. MHGRN transforms the nodes of the graph according to specific types and encodes the knowledge graph using a graph neural network-based method called multi-hop message passing. The previous methods that extract a subgraph for specific entities from an external knowledge graph have a problem that it could bring concepts that are semantically unrelated to context [6].

The question answering graph neural network (QA-GNN) adds embedding of question and choice text encoded by a language model to the subgraph as a node and performs updates to nodes on the subgraph through a graph neural network. And it is used in the process of calculating the relevance score between a question and a choice [16].

Wang et al. proposed a method that trains a path generator predicting the paths between entities in a question and a choice in a knowledge graph using a pre-trained language model. It uses the path generator to solve the problem where the path may not exist between the two entities that have a semantic relation [17].

The align mask and select (AMS) was proposed based on the fact that BERT can capture knowledge contained in the text during pre-training. AMS automatically generates pseudo-problems similar to CommonsenseQA dataset based on the content of ConceptNet and uses them for fine-tuning BERT along with CommonsenseQA dataset [18]. There is no information about the context in which concepts are used in a general knowledge graph. To solve this problem, descriptive knowledge for commonsense question answering (DEKCOR) brings a description of the entities that appear in questions and choices from an external dictionary and provides them together to the language model [19].

## 3 Proposed Methodology

### 3.1 Model Structure

The overall structure of the model is shown in Fig. 4. The overall flow of proposed model is as follows. Given a question-and-choice pair, it enters both the text encoder and the knowledge extractor simultaneously. The input that goes into the text encoder is encoded via ALBERT. A knowledge extractor extracts entities and relationships that exist in (Q, A) pairs from knowledge graph. The result created in this way is a schema graph. We expanded the created schema graph using SGE. The graph encoder embeds an expanded schema graph. At this time, we do not know which entity is important, so we use the information created by the text encoder. In Fig. 4, the output of the text encoder is used for two purposes. The output to the right is intended to provide information to the classifier about the content of the text. On the other hand, the output going down is for extracting information relevant to the problem when doing graph encoding. The classifier uses a combination of graph and text embedding to score points. In this study, SGE techniques were applied to the knowledge extractor. We used the latest pre-trained language model, ALBERT, as a text encoder for embedding question text answer example text. Moreover, we utilized KagNet and MHGRN as a graph encoder for embedding schema graphs.

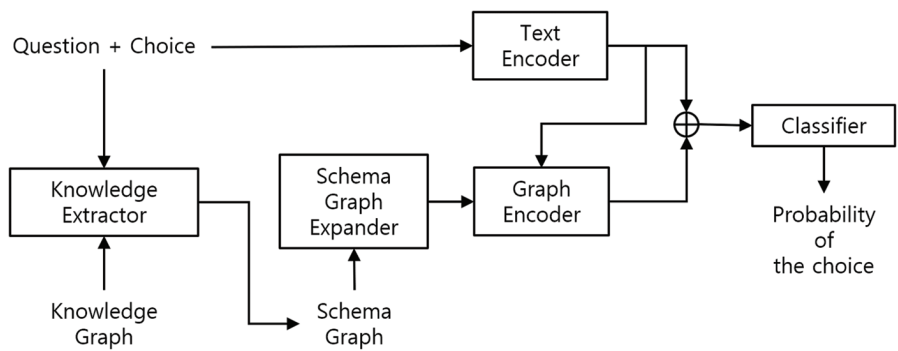

**Figure 4:** Structure of proposed model

In addition, to simplify the problem, instead of choosing the correct one answer among five choices, the model gets the question and a choice as an input. So we turn it into a problem that, given each question-choice pair, outputs the probability that that choice is the correct answer. Fig. 5 shows an example of this. The trained model chooses the choices with the highest score out of five question-choice pairs generated from one problem and five choice examples.

**Figure 5:** Example of conversion of a question

### 3.2 Knowledge Extractor

Given one question-choice pair (Q, A), the knowledge extractor extracts concepts existing in ConceptNet related to (Q, A). We denote the concept set extracted from the question as $C_q$ and the concept set extracted from the choice as $C_a$. During this process, we converted all the text to lowercase letters, tokenized and did lemmatization. After extracting entities from ConceptNet in this way, we found all paths connecting $c_i \in C_q$ and $c_j \in C_a$. If you do not limit the length of the path, there can be thousands of paths between the two entities in ConceptNet. Therefore, from a realistic point of view, we exclude paths that exceed a certain length of K. This creates a new knowledge graph that contains all the concepts and paths obtained for a pair of (Q, A). This is the process of extracting only the subgraphs related to Q and A from ConceptNet. The resulting graph is called the schema graph for (Q, A). Fig. 6 shows an example of such a schema graph. Colored words represent concepts extracted from questions and choices. Green words are concepts related to the choice, and blue words are concepts related to questions. This knowledge extractor concept is already widely used in KagNet, MHGRN, and so on [5,6].

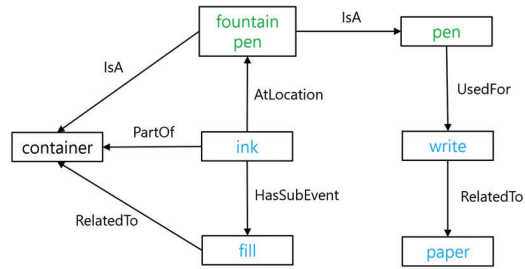

**Figure 6:** Example of schema graph

*3.3 Text Encoder*

To embed the input question-choice pair (Q, A), we use a text encoder. As mentioned in Section 2.2, the pre-trained language model BERT has been very successful. However, BERT took up a large amount of space in the graphic card memory so it was unable to scale further due to resource limitations. To solve this problem of BERT, ALBERT [20] was invented. In general, machine learning (ML) models tend to perform better as they grow in size and use more data, except in cases of overfitting. However, ALBERT, which is a light version of BERT, can achieve similar performance as BERT by using less memory space. This allows you to leverage free space to create higher-performance models. Tab. 1 compares the different structures of BERT and ALBERT. Looking at the GLUE scores in Tab. 1, the performance of the BERT-base model and the ALBERT-large model is similar. However, the number of parameters in ALBERT-large is only 1/6 of the number of parameters in BERT-base [20]. ALBERT shows the best score in CommonsenseQA dataset among recent language representation models which don't use other additional methods [6].

**Table 1:** Characteristics of BERT and ALBERT

| Model | Params | Layers | Hidden Dim. | Embed Dim. | GLUE Score |
| --- | --- | --- | --- | --- | --- |
| BERT-base | 108 M | 12 | 768 | 768 | 82.3 |
| BERT-large | 334 M | 24 | 1024 | 1024 | 85.2 |
| ALBERT-base | 12 M | 12 | 768 | 128 | 80.1 |
| ALBERT-large | 18 M | 24 | 1024 | 128 | 82.4 |
| ALBERT-xlarge | 60 M | 24 | 2048 | 128 | 85.5 |
| ALBERT-xxlarge | 235 M | 12 | 4096 | 128 | 88.7 |

As a text encoder, Lin et al. [5] and Feng et al. [6] used BERT and RoBERTa, respectively. In this study, we applied the more recent model, ALBERT. Language models such as BERT, RoBERTa, and ALBERT take one sentence as input, not a pair of (Q, A). Therefore, we must transform (Q, A) pair to a single sentence. Therefore, we make a declarative sentence by replacing the interrogative word in the question text with a choice text. BERT and ALBERT basically perform word-by-word embedding, not sentence-by-sentence. Thus, we have to insert a special token "[CLS]" representing the embedding of the entire sentence at the beginning of the input sentence [20]. Additionally, we have to insert a special token "[SEP]" meaning the end of the sentence at the end of each sentence. In summary, we obtain vector representations for question-choice pairs by entering text in the form of "[CLS] sentence [SEP]" into ALBERT.

### 3.4 Graph Encoder

A graph encoder is used to obtain the embeddings of nodes that make up the schema graph. In this study, we propose a method using KagNet [5] and MHGRN [6] among previous studies. KagNet uses graph convolution networks to obtain embeddings of nodes. Path embeddings of graphs are obtained using LSTM. Finally, the attention mechanism [21] is used to calculate the embedding of the entire graph. MHGRN works similarly, but the difference is that it works in a way called multi-hop message passing.

### 3.5 Classifier

The classifier uses the output vector of the text encoder and the output vector of the graph encoder to calculate the final score of the question-choice pair fed into the model. This classifier is a one-layer neural network. When the results of the two modules are compressed into a single vector and given as an input to the classifier, it outputs a final output value between 0 and 1 through a sigmoid function. The larger this final output value, the more likely the input question-choice pair is to be correct.

### 3.6 Schema Graph Expander (SGE)

In previous studies, schema graphs were extracted and used immediately without additional processing. In this study, after extracting the schema graph, we expand the schema as additional processing. Existing models extracted schema graphs only for entities that appear in questions and choices. However, in the real world, the concepts that form the 'IsA' relationship share many similar properties. For example, 'dog' and 'puppy' are different words but they are very similar concepts except their age. Motivated by this fact, we extracted an expanded schema graph that uses both original concepts that appear in questions and choices. SGE adds another concept similar to the original concept with an 'IsA' relationship. This is the addition of entities that are 'IsA' relationships to $C_q$'s elements and $C_a$'s elements. Ideally, it's best to leverage all entities that are 'IsA' relationships to every entity. However, as the number of added entities increased, we found experimentally that the amount of graph computation became excessively large. Therefore, in this study, only one of the entities in the 'IsA' relationship is randomly added for each entity.

## 4 Dataset Description

CommonsenseQA dataset [1] was created for model training and model performance evaluation for CSQA. Each question consists of one natural language question and five answer choices for that question. A real example of a question in the dataset is shown in Fig. 7. In the CommonsenseQA dataset, a problem consists of a question asking which entities have a specific relationship to an entity and five candidates for choice. Talmor et al. extracted connected concepts from ConceptNet and generated a set of problems containing these concepts through crowdsourcing. A summary of the data set is shown in Tab. 2. There are a total of 12,102 commonsense questions and they are divided into training set, validation set and test set. Since the test set's answers are not disclosed, the performance evaluation of the model was conducted using the official validation set in this study.

## 5 Experiments

All codes were written based on Python 3.6 and PyTorch 1.4.0 was used for the deep learning framework. All language models including ALBERT were implemented using Transformer 2.10.0 library. As GPU, GeForce TITAN Xp (VRAM 12GB) was used.

What do people aim to do at work?
A. *complete job*
B. learn from each other
C. kill animals
D. wear hats
E. talk to each other

**Figure 7:** Example of CommonsenseQA dataset question

**Table 2:** Characteristics of CommonsenseQA dataset

| | |
|---|---|
| Total questions | 12,094 |
| Train set | 9,741 |
| Dev set | 1,221 |
| Test set | 1,140 |
| Avg. words in questions | 13.41 |
| Avg. words in answers | 1.5 |

Extracting schema graphs from knowledge graphs is not a part of neural network or training. Therefore, we first construct schema graphs on all the data in CommonsenseQA dataset as preprocessing and in training time we provide the constructed schema graphs and question-choice pairs as inputs to our model. In the case of using KagNet as a text encoder, the model is trained about the dataset independently as shown in [5]. On the other hand, when MHGRN was used as a text encoder, the entire model was trained end-to-end as shown in [6]. Cross entropy was used as a loss function and adaptive moment estimation (Adam) [22] and rectified adaptive moment estimation (RAdam) [23] were used as the optimizer for the model, as shown in [5] and [6], respectively. In addition, we use the early stopping technique which terminates the training early if the highest score was not updated more than 3 times in a row every epoch.

We used ALBERT v2 which hyperparameters are optimized. All sentences were tokenized using the SentencePiece technique [24] used by ALBERT [20]. We experimented by applying KagNet and MHGRN to all sizes (base, large, xlarge, xxlarge) of ALBERT. For comparison the performance was measured by experimenting with the models proposed in [5] and [6] in the same environment. The results are shown in Fig. 8.

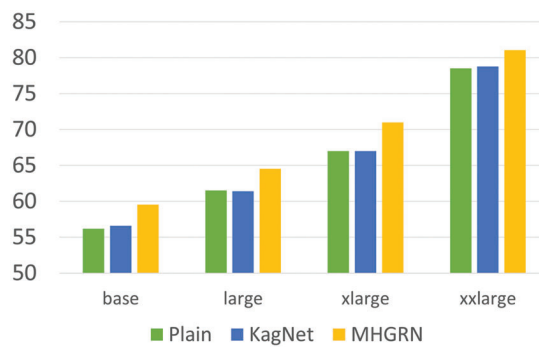

**Figure 8:** Performance comparison according to the size of ALBERT

Even without the use of additional knowledge graphs or modules, the plain xxlarge ALBERT model already shows performance close to the best performance of previous studies. This shows that even in the CSQA problem, transfer learning is very effective just like in other NLP tasks. ALBERT+KagNet models had slight performance improvements when ALBERT size in base or xxlarge. However, the performance improvement was negligible for large and xlarge. For the ALBERT+MHGRN model, there was about 3-point score improvement for all sizes of ALBERT. In particular, the ALBERT-xxlarge+MHGRN model shows better performance than previous studies (Tab. 3). This implies that MHGRN can supplement the knowledge that the ALBERT language model has.

Table 3: Comparison of the best result with existing methods

| Model | Accuracy |
| --- | --- |
| BERT-large+KagNet | 63.55 |
| RoBERTa-large+MHGRN | 77.15 |
| ALBERT-xxlarge+MHGRN | 81.08 |

The performance of BERT+KagNet was evaluated in [5] and RoBERTa+MHGRN was evaluated in [6]. However, the performances were not evaluated for RoBERTa+KagNet or BERT+MHGRN combinations. Therefore, we conducted the experiments as well. The results are shown in Tab. 4. KagNet showed some performance improvements in BERT and ALBERT but in RoBERTa the performace was slightly decreased. We can infer that the RoBERTa model already knows the knowledge provided by KagNet, but rather overfits due to KagNet. MHGRN has performance improvements for all models BERT, RoBERTa and ALBERT.

Table 4: Accuracy of applying KagNet and MHGRN

|  | Plain | KagNet | MHGRN |
| --- | --- | --- | --- |
| BERT-large | 61.83 | 63.55 | 65.27 |
| RoBERTa-large | 75.92 | 75.51 | 77.15 |
| ALBERT-xxlarge | **78.54** | **78.79** | **81.08** |

Tab. 5 and Fig. 9 show the results of applying the schema graph expansion (SGE) to the KagNet and MHGRN models and applying KagNet-SGE and MHGRN-SGE to the language models. For MHGRN, there was a score improvement of 1.4 point for RoBERTa-large and about 1 point for ALBERT-base and ALBERT-large. From this, it can be seen that the schema graph expansion method is effective to some extent. However, since it is not a consistent performance improvement, the schema graph expansion method needs to be more sophisticated and this is our future study.

Table 5: Accuracy of SGE on KagNet and MHGRN

|  | KagNet | KagNet-SGE | MHGRN | MHGRN-SGE |
| --- | --- | --- | --- | --- |
| BERT-large | 63.55 | 62.74 | **65.27** | 64.37 |
| RoBERTa-large | 75.51 | 76.17 | 77.15 | **78.54** |
| ALBERT-base | 56.59 | 56.43 | 59.54 | **60.52** |
| ALBERT-large | 61.43 | 61.43 | 64.54 | **65.27** |
| ALBERT-xlarge | 66.99 | 67.24 | **71.01** | 70.76 |
| ALBERT-xxlarge | 78.79 | 78.87 | **81.08** | 79.93 |

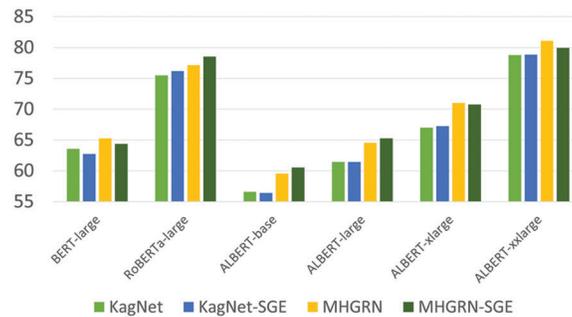

**Figure 9:** Comparison of schema graph expansion on KagNet and MHGRN

## 6 Conclusion

Previous studies which is utilizing knowledge graphs for CSQA have only used relatively older pre-trained language models such as BERT or RoBERTa models. In this study, we propose a model to apply external knowledge graphs to the latest language model ALBERT to solve the CSQA problem. Therefore, our study can be viewed as an update of the old pre-trained model of the existing study for CSQA to the latest pre-trained language model. Through various experiments, we confirm our model performs better than previous studies such as KagNet or MHGRN. We also try a new novel approach called schema graph expansion to reflect the relationship between entities which are having 'IsA' relationship. This Schema graph expansion can be viewed as utilizing a synonym relationship. Although schema graph expansions provide some performance gains but it was not consistent performance gains for all structure and size of the language model. Therefore, it will be interesting further research to figure out this cause. As a future study, we will also develop better schema graph expansion because there is still room for performance improvement. The schema graph expansion of this study expanded only one word which is an ISA relationship. We will expand on multiple words that are ISA relationships for future study. Also, when we expand the graph, considering the context of the word in the sentences will be an interesting future research topic.

**Funding Statement:** This work was supported by the National Research Foundation of Korea (NRF) grant funded by the Korea Government (MSIT) (No.2020R1G1A1100493).

**Conflicts of Interest:** The authors declare that they have no conflicts of interest to report regarding the present study.